\documentclass[10pt]{article}

\usepackage[margin=1in]{geometry}
\usepackage{times}
\usepackage{microtype}
\usepackage{graphicx}
\usepackage{booktabs}
\usepackage{multirow}
\usepackage{amsmath,amssymb}
\usepackage{url}
\usepackage{xcolor}
\usepackage{enumitem}
\usepackage{caption}
\usepackage{subcaption}

\usepackage[numbers,sort&compress]{natbib}

\usepackage{hyperref}
\hypersetup{
    colorlinks=true,
    linkcolor=blue,
    citecolor=blue,
    urlcolor=blue
}

\title{SVF-CR: Synchronized Visual-Facial Cross-Refinement for Multimodal Ambivalence and Hesitancy Recognition}

\author{
Hyein Park$^{1}$ \quad Namho Kim$^{2}$ \quad Junhwa Kim$^{1,*}$\\
$^{1}$Dept. of AI Software Convergence, Konyang University, $^{2}$Korean Broadcasting System (KBS)\\
\texttt{26810503@konyang.ac.kr, namho96@kbs.co.kr, junhwakim@konyang.ac.kr}\\
$^{*}$Corresponding author
}

\date{}

\begin{document}
\maketitle

\begin{abstract}
Ambivalence and hesitancy are subtle behavioral states that are expressed through a combination of verbal content, facial behavior, visual context, and acoustic cues. Effective recognition therefore requires not only extracting informative unimodal representations, but also modeling how temporally aligned behavioral evidence interacts across modalities. In this paper, we propose a synchronized visual-facial cross-refinement framework (SVF-CR) with pairwise multimodal evidence fusion for ambivalence and hesitancy recognition. The proposed method first extracts whole-video segment tokens and cropped-face segment tokens using the same temporal partition. The synchronized visual and facial tokens are refined through intra-modal self-attention and bidirectional visual-facial cross-attention, allowing whole-video context and local facial behavior to mutually refine each other before evidence construction. We then construct segment-level visual-facial evidence using consistency and discrepancy modeling, followed by temporal self-attention and attention pooling. Textual and acoustic features are lightly refined through context self-attention and are fused with the enhanced visual-facial evidence at the final decision stage using pairwise evidence fusion. Experiments on the BAH (Behavioral Ambivalence/Hesitancy) public evaluation split show that the proposed synchronized visual-facial cross-refinement improves public macro-F1 over both global visual-face token fusion and synchronized evidence baselines, achieving a public macro-F1 of 0.7156. Code is available at : https://github.com/hiinnnii/BAH-Challenge-ECCV2026\_SVF-CR.
\end{abstract}

\section{Introduction}

Understanding human ambivalence and hesitancy is important in many real-world interaction scenarios, including healthcare counseling, behavioral intervention, education, customer interviews, and human--computer interaction \cite{gonzalez2026multimodal, hall1995nonverbal}. In such settings, a person may not explicitly state uncertainty or reluctance, but may still reveal it through subtle verbal and non-verbal behaviors. Automatically recognizing these states can help interactive systems identify when a user is uncertain, resistant, or not fully committed, and can support more adaptive feedback, decision support, or intervention strategies \cite{hayashi2023methods}. Therefore, ambivalence and hesitancy recognition is not merely a binary classification problem, but a step toward understanding complex human intention and readiness for change in natural interactions \cite{davidson2020understanding}.

Ambivalence and hesitancy are challenging because they are often expressed through weak, indirect, and temporally distributed cues. Unlike conventional emotion categories, these states are not always conveyed by a single clear facial expression or a specific spoken phrase \cite{kollias2023multi, gonzalez2025bah}. A participant may verbally provide an answer while showing uncertainty through facial movements, gaze changes, pauses, prosodic variation, or inconsistent non-verbal behavior \cite{hayashi2023methods, krahmer2005children, pantic2003toward}. In other cases, the verbal content may appear neutral, while facial and visual behavior reveals hesitation. This makes ambivalence and hesitancy recognition highly dependent on the relationship among textual, visual, facial, and acoustic evidence rather than on any single modality alone.

A common strategy in affective behavior analysis is to combine multiple modalities, such as text, visual, and audio features \cite{zhao2021emotion, kim2025aligning}. However, simple concatenation of heterogeneous modality features does not necessarily lead to better recognition. In the BAH (Behavioral Ambivalence/Hesitancy) task \cite{gonzalez2025bah}, labels are assigned at the video level, while different modalities provide evidence at different temporal and semantic resolutions. Textual features often encode high-level semantic information related to uncertainty or decision making, acoustic features may capture hesitancy-related speaking patterns, and visual observations contain behavioral cues that may not be explicitly expressed in language. Among visual observations, facial behavior is particularly important, but cropped-face features should not be treated simply as an independent global modality. Rather, facial cues are local visual evidence that should be interpreted together with the whole-video context.

Motivated by this observation, we propose a synchronized visual-facial cross-refinement framework for ambivalence and hesitancy recognition. Instead of representing the whole video using only a single global visual vector, we divide the video into multiple temporal segments and extract whole-video segment tokens. In parallel, cropped-face observations are represented as facial segment tokens using the same temporal partition. This temporal synchronization enables the model to compare whole-video context and facial behavior at the segment level. As a result, the model can construct visual-facial evidence from aligned behavioral cues rather than relying on a global visual summary or treating face features as a separate modality.

The core component of the proposed method is a synchronized visual-facial cross-refinement module. Given whole-video segment tokens and facial segment tokens, we first apply intra-modal self-attention to each stream. This allows video tokens to model temporal relations among whole-video segments, and facial tokens to model temporal relations among facial segments. We then apply bidirectional visual-facial cross-attention, where video tokens attend to facial tokens and facial tokens attend to video tokens. Through this process, whole-video context and local facial behavior mutually refine each other before evidence construction. This design is more expressive than shallow feature concatenation because synchronized visual and facial tokens interact before being summarized.

After cross-refinement, we construct segment-level visual-facial evidence. For each synchronized segment, the refined video token and refined facial token are combined using concatenation, element-wise product, and absolute difference. This evidence construction captures both consistency and discrepancy between whole-video context and facial behavior. The resulting evidence tokens are further processed by temporal self-attention, allowing the model to compare ambivalence- and hesitancy-related cues across segments. Finally, attention pooling converts the refined evidence sequence into an enhanced visual-facial representation.

Textual and acoustic evidence are incorporated at the final evidence fusion stage. In our main model, text and audio features are first refined through a lightweight context self-attention layer. However, they are not directly injected into the intermediate visual-facial evidence construction stage. Our experiments indicate that introducing text/audio context into visual-facial refinement can introduce noisy interactions and reduce public macro-F1. Instead, textual evidence, enhanced visual-facial evidence, and acoustic evidence are fused after visual-facial refinement using pairwise evidence fusion. For each pair of evidence vectors, we model their concatenation, element-wise product, and absolute difference, enabling the classifier to capture both agreement and mismatch among modalities.

The main contributions of this work are summarized as follows:
\begin{itemize}[leftmargin=*]
\item We propose a synchronized visual-facial cross-refinement framework (SVF-CR) for ambivalence and hesitancy recognition, where whole-video segment tokens and cropped-face segment tokens are temporally aligned before evidence construction.
\item We introduce bidirectional visual-facial cross-attention to allow whole-video context and facial behavioral cues to mutually refine each other at the segment level.
\item We construct segment-level visual-facial evidence using consistency and discrepancy modeling, followed by temporal self-attention and attention pooling.
\item We fuse textual, enhanced visual-facial, and acoustic evidence through pairwise evidence fusion, capturing both agreement and mismatch among modalities.
\item Experiments on the BAH public evaluation split show that synchronized visual-facial evidence improves over global visual-face token fusion, and the proposed cross-refinement module further improves public macro-F1 to 0.7156.
\end{itemize}
\section{Related Work}

\paragraph{Multimodal affective behavior analysis.}
Affective behavior analysis in the wild aims to recognize human states from unconstrained videos by integrating multiple cues, including language, facial expressions, visual behavior, and speech \cite{zhang2016multimodal}. Since real-world videos often contain noisy or missing modality information, recent multimodal approaches focus on learning complementary evidence rather than relying on a single dominant modality \cite{ma2022multimodal}.

\paragraph{Ambivalence and hesitancy recognition.}
Ambivalence and hesitancy recognition is challenging because the target state is often expressed implicitly rather than through a single obvious emotional signal \cite{song2021hidden}. Prior work has approached this problem from different angles, including linguistic uncertainty modeling, facial behavior analysis, speech prosody analysis, body movement analysis, and cross-modal disagreement modeling \cite{luo2025triagedmsa}. Compared to these largely unimodal or late-fusion approaches, our work explicitly models temporally synchronized visual-facial interaction as an intermediate step before multimodal fusion, rather than treating each modality independently until the final decision stage.

\paragraph{Visual and facial representation.}
Whole-video features capture global context, body motion, and scene-level behavioral cues, whereas cropped-face features provide more localized facial information such as gaze, expression, and subtle facial changes \cite{zheng2022general}. However, treating face features as a completely independent modality can introduce redundancy and noise, especially when the dataset is limited. In this work, we use synchronized face tokens to refine whole-video visual tokens, allowing local facial behavior and global visual context to interact.

\paragraph{Cross-attention and pairwise fusion.}
Cross-attention has been widely used in multimodal learning to allow one representation to selectively attend to another. In addition, pairwise interaction features, such as concatenation, element-wise product, and absolute difference, are commonly used to model agreement and discrepancy between representations \cite{chen2021crossvit}. Inspired by these ideas, our SVF-CR framework performs bidirectional visual--facial cross-refinement and constructs explicit pairwise evidence for final multimodal fusion.
\section{Proposed Method}
\label{sec:method}

\begin{figure*}[t]
\centering
\includegraphics[width=\textwidth,height=10cm]{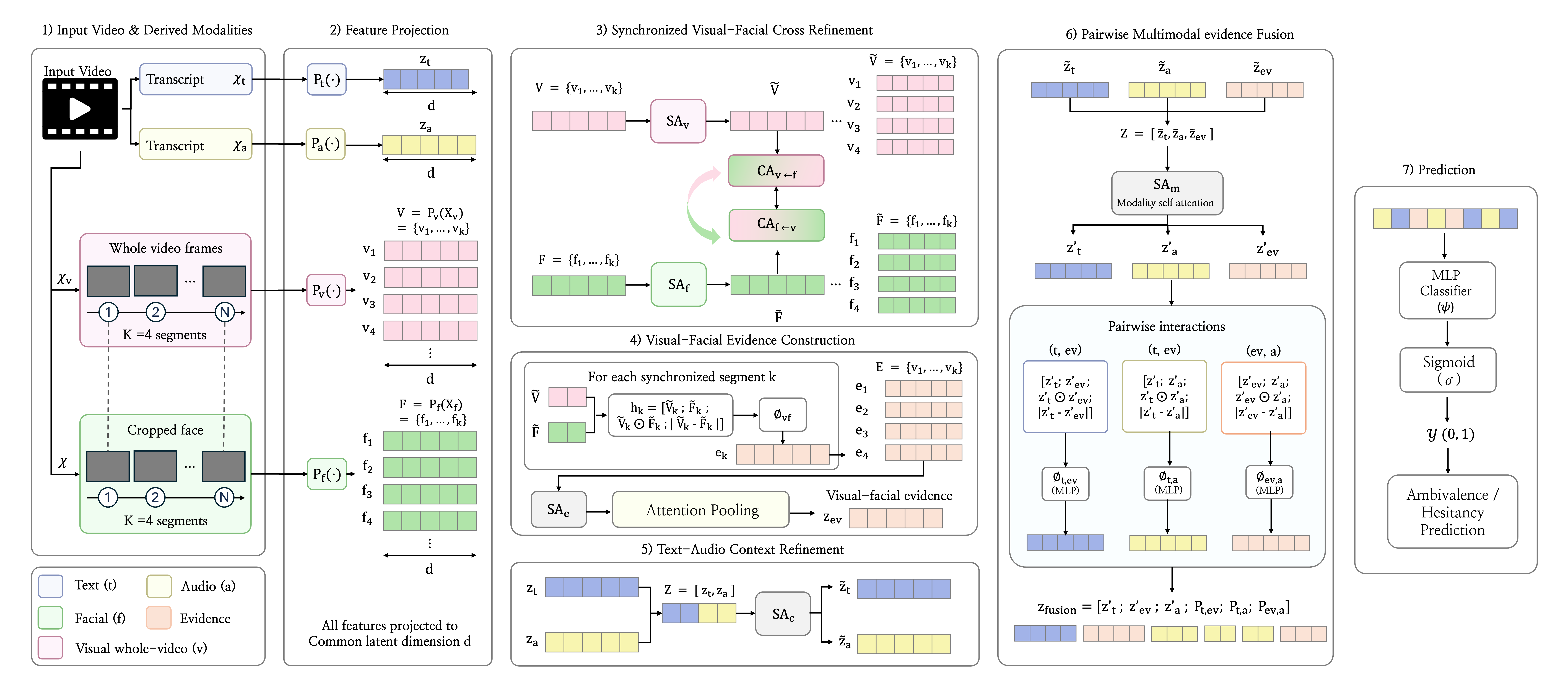}
\caption{Overview of the proposed Synchronized Visual-Facial Cross-Refinement (SVF-CR) framework. The model extracts textual, whole-video visual, cropped-face facial, and acoustic features from an input video. Whole-video and facial segment tokens are temporally synchronized and refined through intra-modal self-attention and bidirectional visual-facial cross-attention. The refined tokens are then used to construct visual-facial evidence, which is finally fused with textual and acoustic evidence through pairwise multimodal evidence fusion.}
\label{fig:method_overview}
\end{figure*}

We propose a synchronized visual-facial cross-refinement framework with pairwise multimodal evidence fusion for ambivalence and hesitancy recognition. As illustrated in Fig.~\ref{fig:method_overview}, the proposed method first extracts temporally synchronized whole-video and cropped-face segment tokens. These tokens are refined through intra-modal self-attention and bidirectional visual-facial cross-attention. The refined segment tokens are then converted into visual-facial evidence tokens using consistency and discrepancy modeling. Finally, the enhanced visual-facial evidence is fused with textual and acoustic evidence through pairwise evidence fusion.

\subsection{Overview}

Given an input video, the goal is to predict a binary label
\begin{equation}
y \in \{0,1\},
\end{equation}
where $y=1$ indicates the presence of ambivalence or hesitancy.

For each sample, we extract textual, visual, facial, and acoustic features:
\begin{equation}
\mathbf{x}_t,\quad
\mathbf{X}_v=\{\mathbf{x}_v^1,\ldots,\mathbf{x}_v^K\},\quad
\mathbf{X}_f=\{\mathbf{x}_f^1,\ldots,\mathbf{x}_f^K\},\quad
\mathbf{x}_a .
\end{equation}
Here, $\mathbf{x}_t$ and $\mathbf{x}_a$ denote the textual and acoustic features, respectively. $\mathbf{X}_v$ denotes whole-video segment tokens, and $\mathbf{X}_f$ denotes cropped-face segment tokens. The whole-video and face tokens are extracted using the same temporal partition, so that $\mathbf{x}_v^k$ and $\mathbf{x}_f^k$ correspond to the same temporal segment. In our implementation, we use $K=4$.

All features are projected into a common latent dimension $d$:
\begin{equation}
\mathbf{z}_t=P_t(\mathbf{x}_t),\quad
\mathbf{z}_a=P_a(\mathbf{x}_a),
\end{equation}
\begin{equation}
\mathbf{V}=P_v(\mathbf{X}_v),\quad
\mathbf{F}=P_f(\mathbf{X}_f),
\end{equation}
where $\mathbf{V}=\{\mathbf{v}_1,\ldots,\mathbf{v}_K\}$ and $\mathbf{F}=\{\mathbf{f}_1,\ldots,\mathbf{f}_K\}$ are the projected whole-video and facial segment token sequences.

\subsection{Synchronized Visual-Facial Cross-Refinement}

The core component of our method is synchronized visual-facial cross-refinement. We first refine each token sequence using intra-modal self-attention:
\begin{equation}
\widetilde{\mathbf{V}}=\mathrm{SA}_v(\mathbf{V}),\quad
\widetilde{\mathbf{F}}=\mathrm{SA}_f(\mathbf{F}).
\end{equation}
This allows whole-video tokens and facial tokens to model temporal relations within their own streams.

We then apply bidirectional cross-attention between the two synchronized streams:
\begin{equation}
\widehat{\mathbf{V}}
=
\mathrm{CA}_{v\leftarrow f}
\left(
\widetilde{\mathbf{V}},
\widetilde{\mathbf{F}}
\right),
\end{equation}
\begin{equation}
\widehat{\mathbf{F}}
=
\mathrm{CA}_{f\leftarrow v}
\left(
\widetilde{\mathbf{F}},
\widetilde{\mathbf{V}}
\right).
\end{equation}
In this process, whole-video segment tokens attend to facial segment tokens, and facial segment tokens attend to whole-video segment tokens. This enables global visual context and local facial behavior to mutually refine each other before evidence construction.

Each self-attention or cross-attention block consists of multi-head attention, residual connection, layer normalization, and a feed-forward residual block.

\subsection{Visual-Facial Evidence Construction}

After cross-refinement, we construct segment-level visual-facial evidence. For each synchronized segment $k$, the refined whole-video token $\widehat{\mathbf{v}}_k$ and facial token $\widehat{\mathbf{f}}_k$ are combined as
\begin{equation}
\mathbf{h}_k =
\left[
\widehat{\mathbf{v}}_k ;
\widehat{\mathbf{f}}_k ;
\widehat{\mathbf{v}}_k \odot \widehat{\mathbf{f}}_k ;
\left|
\widehat{\mathbf{v}}_k-\widehat{\mathbf{f}}_k
\right|
\right],
\end{equation}
where $\odot$ denotes element-wise multiplication and $[\cdot;\cdot]$ denotes concatenation. This representation captures both consistency and discrepancy between whole-video context and facial behavior.

The segment-level evidence token is obtained by
\begin{equation}
\mathbf{e}_k=\phi_{vf}(\mathbf{h}_k),
\end{equation}
where $\phi_{vf}(\cdot)$ is a multilayer perceptron. The evidence sequence is
\begin{equation}
\mathbf{E}=\{\mathbf{e}_1,\ldots,\mathbf{e}_K\}.
\end{equation}

To model temporal relations among evidence tokens, we apply evidence self-attention:
\begin{equation}
\widetilde{\mathbf{E}}=\mathrm{SA}_e(\mathbf{E}).
\end{equation}
The final visual-facial evidence vector is obtained by attention pooling:
\begin{equation}
\alpha_k =
\frac{
\exp(s(\widetilde{\mathbf{e}}_k))
}{
\sum_{j=1}^{K}\exp(s(\widetilde{\mathbf{e}}_j))
},
\quad
\mathbf{z}_{ev}
=
\sum_{k=1}^{K}
\alpha_k \widetilde{\mathbf{e}}_k ,
\end{equation}
where $s(\cdot)$ is a learnable scoring function.

\subsection{Text-Audio Context Refinement and Pairwise Evidence Fusion}

Textual and acoustic features provide complementary semantic and prosodic evidence for affective and behavioral understanding. In our main model, these features are not injected into the intermediate visual-facial evidence construction stage. Instead, we first refine text and audio as a two-token context sequence using a lightweight self-attention layer:
\begin{equation}
\mathbf{C}
=
[\mathbf{z}_t, \mathbf{z}_a]
\in \mathbb{R}^{2 \times d},
\end{equation}
\begin{equation}
\widetilde{\mathbf{C}}
=
\mathrm{SA}_{c}(\mathbf{C})
=
[\widetilde{\mathbf{z}}_t, \widetilde{\mathbf{z}}_a],
\end{equation}
where $\widetilde{\mathbf{z}}_t$ and $\widetilde{\mathbf{z}}_a$ denote the refined textual and acoustic evidence vectors, respectively.

Unlike the contextual refinement variant evaluated in our ablation study (Section 4.3), the main model does not inject textual or acoustic evidence into the intermediate visual-facial evidence construction stage. Instead, textual evidence, enhanced visual-facial evidence, and acoustic evidence are fused at the final decision stage. This design keeps the visual-facial evidence construction focused on synchronized behavioral cues, while allowing text and audio to contribute through final multimodal evidence fusion.

The final evidence tokens are defined as
\begin{equation}
\mathbf{Z}
=
[
\widetilde{\mathbf{z}}_t,
\mathbf{z}_{ev},
\widetilde{\mathbf{z}}_a
]
\in \mathbb{R}^{3 \times d}.
\end{equation}
Before constructing pairwise evidence, we further apply modality-level self-attention over the three evidence tokens :
\begin{equation}
\mathbf{Z}'
=
\mathrm{SA}_{m}(\mathbf{Z})
=
[
\mathbf{z}'_t,
\mathbf{z}'_{ev},
\mathbf{z}'_a
].
\end{equation}
This allows textual, visual-facial, and acoustic evidence to interact at the modality level before pairwise evidence construction.

For each pair of evidence vectors $(\mathbf{z}'_i,\mathbf{z}'_j)$, we construct a pairwise evidence representation :
\begin{equation}
\mathbf{p}_{ij}
=
\phi_{ij}
\left(
\left[
\mathbf{z}'_i ;
\mathbf{z}'_j ;
\mathbf{z}'_i \odot \mathbf{z}'_j ;
\left|
\mathbf{z}'_i-\mathbf{z}'_j
\right|
\right]
\right),
\end{equation}
where $\phi_{ij}(\cdot)$ denotes a pair-specific multilayer perceptron. The element-wise product captures agreement between two evidence vectors, while the absolute difference captures their discrepancy.

The final fused representation is obtained by concatenating the modality-level evidence tokens and their pairwise evidence vectors:
\begin{equation}
\mathbf{z}_{fusion}
=
\left[
\mathbf{z}'_t ;
\mathbf{z}'_{ev} ;
\mathbf{z}'_a ;
\mathbf{p}_{t,ev} ;
\mathbf{p}_{t,a} ;
\mathbf{p}_{ev,a}
\right].
\end{equation}
The prediction is computed as
\begin{equation}
\hat{y}
=
\sigma
\left(
\psi(\mathbf{z}_{fusion})
\right),
\end{equation}
where $\psi(\cdot)$ is the final classifier and $\sigma(\cdot)$ is the sigmoid function.

\subsection{Training Objective}

The model is trained with binary cross-entropy loss:
\begin{equation}
\mathcal{L}_{main}
=
\mathrm{BCE}(y,\hat{y}).
\end{equation}

We also attach auxiliary classifiers to the pairwise evidence vectors to encourage each pairwise interaction to be discriminative. Let $\hat{y}_{ij}$ be the auxiliary prediction from pairwise evidence $\mathbf{p}_{ij}$. The auxiliary loss is
\begin{equation}
\mathcal{L}_{aux}
=
\frac{1}{|\mathcal{P}|}
\sum_{(i,j)\in\mathcal{P}}
\mathrm{BCE}(y,\hat{y}_{ij}),
\end{equation}
where $\mathcal{P}$ is the set of modality pairs.

The final objective is
\begin{equation}
\mathcal{L}
=
\mathcal{L}_{main}
+
\lambda
\mathcal{L}_{aux}.
\end{equation}
In our final model, we set $\lambda=0.01$ to regularize pairwise evidence learning without overwhelming the main objective.
\section{Experiments}
\label{sec:experiments}

\subsection{Experimental Setup}

We evaluate the proposed Synchronized Visual-Facial Cross-Refinement (SVF-CR) framework on the BAH binary ambivalence/hesitancy recognition task \cite{gonzalez2025bah, abaw}. The task is formulated as a video-level binary classification problem, where each video is classified as either ambivalent/hesitant or non-ambivalent/non-hesitant.

For model development, we combine the available training and validation data and construct a five-fold train--validation protocol. In each fold, the model is trained on four folds and selected using the remaining fold based on validation macro F1. Public evaluation is performed by averaging the predicted probabilities from the five fold-specific models. Unless otherwise stated, all public results are reported using this five-fold ensemble setting.

We report accuracy (ACC), balanced accuracy (BACC), macro F1 (MF1), area under the ROC curve (AUC), and area under the precision-recall curve (AUPRC). Since the task is binary and the class distribution can be imbalanced, macro F1 and balanced accuracy are used as the main threshold-based metrics. AUC and AUPRC are additionally reported to evaluate the ranking quality of predicted probabilities.

\subsection{Implementation Details}

We extract four types of features: textual, whole-video visual, cropped-face visual, and acoustic features. For textual evidence, each transcript is encoded using a pretrained text embedding model \cite{chen2017esim} and concatenated with hesitation-related cue statistics, resulting in a 1045-dimensional feature. For whole-video visual evidence, each video is divided into $K=4$ temporal segments and encoded using a pretrained Qwen-VL model \cite{qwen}, producing visual segment tokens of size $K \times 3584$. For facial evidence, cropped-face frames are divided using the same temporal partition and encoded using VideoMAE \cite{tong2022videomae}, resulting in facial segment tokens of size $K \times 768$. For acoustic evidence, we use a pretrained speech model \cite{radford2023whisper} to extract a 3072-dimensional audio representation. All features are standardized using statistics fitted only on the training portion of each fold.

All experiments are conducted on Ubuntu 20.04 with 64GB RAM and a single NVIDIA GeForce RTX 4090 GPU. The model is implemented in PyTorch. All modality features are projected into a common latent dimension of 128, and the hidden dimension of the fusion modules is set to 256. The final model uses a dropout rate of 0.4 and an auxiliary pairwise loss weight of 0.01.

Each fold-specific model is trained using AdamW with a learning rate of $1 \times 10^{-4}$ and weight decay of $1 \times 10^{-3}$. The training batch size is fixed to 32. Each model is trained for up to 200 epochs with early stopping based on validation macro F1, with a patience of 40 epochs. During inference, the batch size can be increased because it does not affect the trained model parameters.

\subsection{Results and Ablation Study}

Table~\ref{tab:public_main} shows the main public evaluation results. The global token-cross baseline uses text, enhanced visual evidence, and audio, but does not explicitly construct synchronized visual-facial evidence. The proposed SVF-CR achieves a public macro F1 of 0.7156, improving over the global token-cross baseline. It also achieves higher balanced accuracy, AUC, and AUPRC, indicating better class-balanced prediction and ranking quality.

\begin{table}[t]
\centering
\caption{Main public evaluation results.}
\label{tab:public_main}
\small
\setlength{\tabcolsep}{4pt}
\begin{tabular}{lccccc}
\toprule
Method & ACC & BACC & MF1 & AUC & AUPRC \\
\midrule
Global token-cross & 0.7219 & 0.7097 & 0.7094 & 0.7660 & 0.8336 \\
\textbf{SVF-CR} & \textbf{0.7257} & \textbf{0.7179} & \textbf{0.7156} & \textbf{0.7794} & \textbf{0.8387} \\
\bottomrule
\end{tabular}
\end{table}

We next analyze the effect of modality combinations in Table~\ref{tab:modality_ablation}. The results show that text and audio are strong standalone cues, while enhanced visual evidence alone is relatively weak. However, when synchronized visual-facial evidence is combined with text and audio, the final model achieves the best macro F1 and balanced accuracy. This suggests that visual-facial evidence is most useful as complementary behavioral evidence rather than as a standalone cue.

\begin{table}[t]
\centering
\caption{Effect of modality combinations using the proposed SVF-CR framework.}
\label{tab:modality_ablation}
\small
\setlength{\tabcolsep}{3.5pt}
\begin{tabular}{lcccc}
\toprule
Modalities & BACC & MF1 & AUC & AUPRC \\
\midrule
Text & 0.6867 & 0.6923 & 0.7779 & \textbf{0.8462} \\
eVisual & 0.5861 & 0.5817 & 0.5978 & 0.6772 \\
Audio & 0.6972 & 0.7005 & 0.7690 & 0.8387 \\
Text + eVisual & 0.6913 & 0.6919 & 0.7706 & 0.8437 \\
Text + Audio & 0.7012 & 0.7076 & \textbf{0.7813} & 0.8461 \\
eVisual + Audio & 0.7032 & 0.7039 & 0.7676 & 0.8395 \\
\textbf{Text + eVisual + Audio} & \textbf{0.7179} & \textbf{0.7156} & 0.7794 & 0.8387 \\
\bottomrule
\end{tabular}
\end{table}

Table~\ref{tab:module_ablation} reports the ablation results of synchronized visual-facial evidence modeling strategies. Visual pooling and face pooling evaluate whether whole-video or cropped-face evidence alone is sufficient at the visual-facial evidence stage. Anchor-based token fusion refers to a baseline that uses one modality representation as an anchor and updates it using synchronized visual-facial information, without explicitly constructing segment-wise consistency and discrepancy evidence. The synchronized consistency-discrepancy evidence baseline constructs segment-level visual-facial evidence from aligned video and face tokens using concatenation, element-wise product, and absolute difference, but does not perform bidirectional cross-refinement before evidence construction.

The proposed SVF-CR further applies intra-modal refinement, bidirectional visual-facial cross-attention, segment-level consistency-discrepancy evidence construction, evidence self-attention, and final pairwise evidence fusion. The contextual SVF-CR variant injects text/audio context into the intermediate visual-facial evidence construction stage, while the proposed final model keeps text/audio evidence separate until the final fusion stage.

The results show that the proposed SVF-CR achieves the best macro F1 and balanced accuracy among all module variants. Removing visual-facial cross-attention or using one-directional refinement degrades performance, indicating the importance of bidirectional interaction between whole-video and facial segment tokens. The contextual SVF-CR variant also performs worse than the proposed model, supporting our design choice of keeping visual-facial evidence construction focused on synchronized behavioral cues and fusing text/audio at the final evidence level.

\begin{table}[t]
\centering
\caption{Ablation of synchronized visual-facial evidence modeling modules. All variants use text, enhanced visual evidence, and audio.}
\label{tab:module_ablation}
\small
\setlength{\tabcolsep}{2.8pt}
\begin{tabular}{lcccc}
\toprule
Variant & BACC & MF1 & AUC & AUPRC \\
\midrule
Visual pooling & 0.6848 & 0.6796 & 0.7663 & 0.8313 \\
Face pooling & 0.6955 & 0.6992 & 0.7757 & \textbf{0.8476} \\
Anchor-based fusion & 0.6961 & 0.6799 & 0.7595 & 0.8256 \\
Text-conditioned anchor fusion & 0.6883 & 0.6800 & 0.7635 & 0.8291 \\
Text-conditioned consistency-discrepancy evidence & 0.6878 & 0.6902 & 0.7630 & 0.8329 \\
Sync. consistency-discrepancy evidence & 0.7106 & 0.7099 & 0.7802 & 0.8463 \\
SVF-CR w/o VF cross-attention & 0.6926 & 0.6980 & \textbf{0.7825} & 0.8436 \\
SVF-CR w/o reverse refinement & 0.6789 & 0.6814 & 0.7617 & 0.8336 \\
SVF-CR with text pooling & 0.6932 & 0.6964 & 0.7770 & 0.8417 \\
Contextual SVF-CR & 0.6955 & 0.6944 & 0.7598 & 0.8287 \\
\textbf{SVF-CR} & \textbf{0.7179} & \textbf{0.7156} & 0.7794 & 0.8387 \\
\bottomrule
\end{tabular}
\end{table}

\subsection{Analysis and Discussion}

The experimental results provide several observations. First, the proposed SVF-CR improves over the global token-cross baseline, showing that synchronized visual-facial evidence construction is more effective than using facial tokens only as a global visual enhancement cue. Second, the modality ablation shows that text and audio are strong cues, but the best macro F1 and balanced accuracy are obtained only when synchronized visual-facial evidence is added to them. This indicates that visual-facial evidence contributes complementary behavioral information that improves class-balanced prediction.

Third, the module ablation confirms the importance of bidirectional visual-facial cross-refinement. The synchronized CDI baseline already improves over simpler pooling-based variants, but SVF-CR further improves macro F1 by allowing whole-video and facial segment tokens to mutually refine each other before evidence construction. In contrast, the contextual variant that injects text/audio into intermediate visual-facial evidence construction performs worse, suggesting that text and audio are more effective when fused at the final pairwise evidence fusion stage.

We also examined training sensitivity. The final setting with batch size 32 and learning rate $1 \times 10^{-4}$ achieved the best public macro F1. Reducing the batch size to 16 decreased macro F1 to 0.6981, while increasing it to 64 resulted in 0.7086. Lowering the learning rate to $5 \times 10^{-5}$ and $7.5 \times 10^{-5}$ also reduced macro F1 to 0.7092 and 0.6900, respectively. These results indicate that careful optimization and early stopping are important for this relatively small behavioral recognition task.

Finally, using the saved five-fold ensemble probabilities, we performed an additional dense threshold search on the public evaluation split. The original experimental runner reports a public macro F1 of 0.7156 with a threshold of 0.38. With the same ensemble probabilities, a denser threshold search obtains a public macro F1 of 0.7161 with a threshold of 0.311. Since AUC and AUPRC remain identical, this difference is due only to threshold selection resolution. For consistency with the main experimental runner, we report 0.7156 as the primary public macro F1.

Overall, these findings support the design of SVF-CR, where synchronized visual-facial cues are first refined as behavioral evidence and then fused with textual and acoustic evidence for final ambivalence and hesitancy recognition.

\section{Conclusion}

In this paper, we proposed SVF-CR, a synchronized visual-facial cross-refinement framework for ambivalence and hesitancy recognition. The proposed method refines temporally aligned whole-video and cropped-face segment tokens through intra-modal self-attention and bidirectional cross-attention, and constructs visual-facial evidence using consistency and discrepancy modeling. This evidence is then fused with textual and acoustic cues through pairwise multimodal evidence fusion.

Experimental results on the BAH task show that SVF-CR improves over the global token-cross baseline, achieving a public macro F1 score of 0.7156. Ablation studies further confirm that synchronized visual-facial evidence provides complementary behavioral information and that bidirectional visual-facial refinement contributes to improved class-balanced prediction. Future work will investigate more robust temporal modeling and adaptive fusion strategies for noisy real-world videos.

\bibliographystyle{unsrtnat}
\bibliography{references}

@article{tong2022videomae,
  title={VideoMAE: Masked Autoencoders are Data-Efficient Learners for Self-Supervised Video Pre-Training},
  author={Tong, Zhan and Song, Yibing and Wang, Jue and Wang, Limin},
  journal={Advances in Neural Information Processing Systems},
  year={2022}
}

@inproceedings{radford2023whisper,
  title={Robust Speech Recognition via Large-Scale Weak Supervision},
  author={Radford, Alec and Kim, Jong Wook and Xu, Tao and Brockman, Greg and McLeavey, Christine and Sutskever, Ilya},
  booktitle={International Conference on Machine Learning},
  year={2023}
}

@inproceedings{kim2025aligning,
  title={Aligning Multimodal Data for Fine-Grained Video Understanding via Cross-Attentive Recurrent Fusion},
  author={Kim, Nam-Ho and Kim, Jun-Hwa},
  booktitle={Proceedings of the IEEE/CVF International Conference on Computer Vision},
  pages={113--119},
  year={2025}
}

@inproceedings{chen2017esim,
  title={Enhanced LSTM for Natural Language Inference},
  author={Chen, Qian and Zhu, Xiaodan and Ling, Zhen-Hua and Wei, Si and Jiang, Hui and Inkpen, Diana},
  booktitle={Proceedings of the 55th Annual Meeting of the Association for Computational Linguistics},
  year={2017}
}

@misc{qwen,
  title={Qwen Technical Report},
  author={{Qwen Team}},
  year={2023},
  note={Available online}
}

@misc{abaw,
  title={Affective Behavior Analysis in-the-Wild Challenge},
  author={{ABAW Challenge Organizers}},
  year={2026},
  note={Challenge dataset and task description}
}

@article{gonzalez2026multimodal,
  title={Multimodal Ambivalence/Hesitancy Recognition in Videos for Personalized Digital Health Interventions},
  author={Gonz{\'a}lez-Gonz{\'a}lez, Manuela and Belharbi, Soufiane and Zeeshan, Muhammad Osama and Sharafi, Masoumeh and Aslam, Muhammad Haseeb and Sia, Lorenzo and Richet, Nicolas and Pedersoli, Marco and Koerich, Alessandro Lameiras and Bacon, Simon L and others},
  journal={arXiv preprint arXiv:2604.11730},
  year={2026}
}

@article{zhao2021emotion,
  title={Emotion recognition from multiple modalities: Fundamentals and methodologies},
  author={Zhao, Sicheng and Jia, Guoli and Yang, Jufeng and Ding, Guiguang and Keutzer, Kurt},
  journal={IEEE Signal Processing Magazine},
  volume={38},
  number={6},
  pages={59--73},
  year={2021},
  publisher={IEEE}
}

@article{hall1995nonverbal,
  title={Nonverbal behavior in clinician—patient interaction},
  author={Hall, Judith A and Harrigan, Jinni A and Rosenthal, Robert},
  journal={Applied and preventive psychology},
  volume={4},
  number={1},
  pages={21--37},
  year={1995},
  publisher={Elsevier}
}

@article{gonzalez2025bah,
  title={Bah dataset for ambivalence/hesitancy recognition in videos for behavioural change},
  author={Gonz{\'a}lez-Gonz{\'a}lez, Manuela and Belharbi, Soufiane and Zeeshan, Muhammad Osama and Sharafi, Masoumeh and Aslam, Muhammad Haseeb and Pedersoli, Marco and Koerich, Alessandro Lameiras and Bacon, Simon L and Granger, Eric},
  journal={arXiv preprint arXiv:2505.19328},
  volume={3},
  number={9},
  year={2025}
}

@inproceedings{chen2021crossvit,
  title={Crossvit: Cross-attention multi-scale vision transformer for image classification},
  author={Chen, Chun-Fu Richard and Fan, Quanfu and Panda, Rameswar},
  booktitle={Proceedings of the IEEE/CVF international conference on computer vision},
  pages={357--366},
  year={2021}
}

@inproceedings{zheng2022general,
  title={General facial representation learning in a visual-linguistic manner},
  author={Zheng, Yinglin and Yang, Hao and Zhang, Ting and Bao, Jianmin and Chen, Dongdong and Huang, Yangyu and Yuan, Lu and Chen, Dong and Zeng, Ming and Wen, Fang},
  booktitle={Proceedings of the IEEE/CVF conference on computer vision and pattern recognition},
  pages={18697--18709},
  year={2022}
}

@inproceedings{song2021hidden,
  title={Hidden emotion detection using multi-modal signals},
  author={Song, Byung Cheol and Kim, Dae Ha},
  booktitle={Extended abstracts of the 2021 CHI conference on human factors in computing systems},
  pages={1--7},
  year={2021}
}

@article{luo2025triagedmsa,
  title={TriagedMSA: Triaging sentimental disagreement in multimodal sentiment analysis},
  author={Luo, Yuanyi and Liu, Wei and Sun, Qiang and Li, Sirui and Li, Jichunyang and Wu, Rui and Tang, Xianglong},
  journal={IEEE transactions on affective computing},
  volume={16},
  number={3},
  pages={1557--1569},
  year={2025},
  publisher={IEEE}
}

@inproceedings{zhang2016multimodal,
  title={Multimodal spontaneous emotion corpus for human behavior analysis},
  author={Zhang, Zheng and Girard, Jeff M and Wu, Yue and Zhang, Xing and Liu, Peng and Ciftci, Umur and Canavan, Shaun and Reale, Michael and Horowitz, Andy and Yang, Huiyuan and others},
  booktitle={Proceedings of the IEEE conference on computer vision and pattern recognition},
  pages={3438--3446},
  year={2016}
}

@inproceedings{ma2022multimodal,
  title={Are multimodal transformers robust to missing modality?},
  author={Ma, Mengmeng and Ren, Jian and Zhao, Long and Testuggine, Davide and Peng, Xi},
  booktitle={Proceedings of the IEEE/CVF conference on computer vision and pattern recognition},
  pages={18177--18186},
  year={2022}
}

@article{hayashi2023methods,
  title={Methods to assess ambivalence towards food and diet: a scoping review},
  author={Hayashi, Daisuke and Carvalho, Samantha Dalbosco Lins and Ribeiro, Paula Aver Bretanha and Rodrigues, Roberta Cunha Matheus and S{\~a}o-Jo{\~a}o, Tha{\'\i}s Moreira and Lavoie, Kim and Bacon, Simon and Cornelio, Marilia Estevam},
  journal={Journal of Human Nutrition and Dietetics},
  volume={36},
  number={5},
  pages={2010--2025},
  year={2023},
  publisher={Wiley Online Library}
}

@article{davidson2020understanding,
  title={Understanding and predicting health behaviour change: a contemporary view through the lenses of meta-reviews},
  author={Davidson, Karina W and Scholz, Urte},
  journal={Health psychology review},
  volume={14},
  number={1},
  pages={1--5},
  year={2020},
  publisher={Taylor \& Francis}
}

@inproceedings{kollias2023multi,
  title={Multi-label compound expression recognition: C-expr database \& network},
  author={Kollias, Dimitrios},
  booktitle={Proceedings of the IEEE/CVF conference on computer vision and pattern recognition},
  pages={5589--5598},
  year={2023}
}

@article{pantic2003toward,
  title={Toward an affect-sensitive multimodal human-computer interaction},
  author={Pantic, Maja and Rothkrantz, Leon JM},
  journal={Proceedings of the IEEE},
  volume={91},
  number={9},
  pages={1370--1390},
  year={2003},
  publisher={IEEE}
}

@article{krahmer2005children,
  title={How children and adults produce and perceive uncertainty in audiovisual speech},
  author={Krahmer, Emiel and Swerts, Marc},
  journal={Language and speech},
  volume={48},
  number={1},
  pages={29--53},
  year={2005},
  publisher={SAGE Publications Sage UK: London, England}
}

\end{document}